\documentclass[letterpaper, 10 pt, conference]{ieeeconf}

\IEEEoverridecommandlockouts   %
\usepackage{graphicx}
\usepackage{booktabs}
\usepackage{makecell}
\usepackage{multirow}
\usepackage{siunitx}
\usepackage{placeins}
\usepackage{calc}          %
\usepackage{amsmath}
\usepackage{amssymb}
\usepackage{bm}
\usepackage{xcolor}
\usepackage{tikz}
\usetikzlibrary{arrows.meta}
\usetikzlibrary{patterns}
\usepackage{tabularx}
\usepackage{array}

\usepackage{capt-of}           %
\makeatletter\let\NAT@parse\undefined\makeatother
\usepackage[hidelinks]{hyperref}
\usepackage{cleveref}          %

\makeatletter
\let\keywords\@undefined
\let\endkeywords\@undefined
\makeatother
\newcommand{\keywords}[1]{%
  \vskip 0.6\baselineskip%
  {\small\textbf{\textit{Index Terms}}---#1\par}%
  \vskip 0.4\baselineskip}

\makeatletter
\def\@IEEEsectpunct{\ \,}
\makeatother

\title{\LARGE \bf
\textsc{OmniRobotHome}: A Multi-Camera Home Platform \\ for Real-Time Human-Robot Interaction
}

\newif\ifarxiv
\arxivtrue
\ifarxiv
\author{Junyoung Lee$^{1*}$ \quad Inhee Lee$^{1*}$ \quad Kyungwon Cho$^{1*}$ \quad Jeonghwan Kim$^{1}$ \quad Mingi Choi$^{1}$ \quad Sookwan Han$^{1}$ \\
Chaeyeon Lee$^{1}$ \quad Wonjung Woo$^{1}$ \quad Gunhee Kim$^{1}$ \quad Jisoo Kim$^{1}$ \quad Jeonghyeon Na$^{1}$ \quad Hanbyul Joo$^{1,2}$ \\[0.3em]
$^1$Seoul National University \quad $^2$RLWRLD \\[0.2em]
{\tt\small \href{https://junc0ng.github.io/omnirobothome}{\color{magenta}{https://junc0ng.github.io/omnirobothome}}}%
\thanks{$^{*}$Equal contribution.}%
}
\IEEEaftertitletext{\vspace{-1\baselineskip}}
\newcommand{\floattrim}[1]{}
\makeatletter
\let\arxiv@makecaption\@makecaption
\long\def\@makecaption#1#2{\arxiv@makecaption{#1}{#2}%
  \ifx\@captype\@IEEEtablestring\par\nointerlineskip\vspace{2pt}\fi}
\makeatother
\else
\author{Anonymous Author(s)}
\fi

\definecolor{tabfirst}{rgb}{1, 0.7, 0.7} %
\definecolor{tabsecond}{rgb}{1, 0.85, 0.7} %
\definecolor{tabthird}{rgb}{1, 1, 0.7} %

\providecommand{\floattrim}[1]{\vspace{#1}}

\newcommand{\figref}[1]{Fig.~\ref{#1}}
\newcommand{\tabref}[1]{Tab.~\ref{#1}}

\newcommand{\secref}[1]{Sec.~\ref{#1}}

\definecolor{darkgreen}{cmyk}{0.8,0,0.8,0.4}

\makeatletter
\newcommand{\methodreviewred}{}
\newcommand{\experimentreviewred}{}
\makeatother

\begin{document}
\raggedbottom

\maketitle
\thispagestyle{empty}
\pagestyle{empty}

\begin{abstract}
Robots that assist people must respond promptly and appropriately to human activity, which requires low-latency processing, accurate perception, and anticipation of human motion. These requirements are especially difficult to meet in living environments, where perception must cover wide areas, remain robust to occlusion, and capture natural behavior without wearable devices.
We present \textsc{OmniRobotHome}, a testbed that addresses these challenges in a fully furnished home that supports natural everyday activity. Its 48 hardware-synchronized cameras provide room-wide coverage, enabling accurate markerless perception under occlusion. A distributed perception pipeline processes these streams in parallel and delivers persistent human and object states to robot controllers with low latency. We demonstrate \textsc{OmniRobotHome} through two human-robot interaction studies: proactive yielding for safety and nonverbal intent recognition for assistance. Through these studies, we analyze how camera coverage, state latency, multi-view pose, hand, and gaze estimation, and behavior-memory-based forecasting affect overall task performance. Our results highlight the value of \textsc{OmniRobotHome} as a testbed for studying human-robot assistance in real living environments.

\end{abstract}


\section{Introduction}

Enabling robots to assist with or replace human labor in everyday living environments remains a fundamental objective in robotics. 
Unlike structured industrial workspaces or constrained tabletop scenarios, household environments are inherently unstructured and dynamic. 
Humans move freely across the space, interact with diverse objects, and continuously reconfigure their surroundings. 
Robots therefore need to track people and objects
accurately and continuously with low latency
to respond to these changes.
Furthermore, proactive assistance requires robots to anticipate human actions over short time horizons.
The information needed for such prediction,
as well as for interpreting human instructions,
is not confined to the robot's local workspace.
Perception must extend across the living space.

Existing robot systems typically rely on cameras mounted on robots or fixed around their local workspaces~\cite{brohan2023RT1, kalashnikov2018QTOpt, shridhar2022CLIPort, zeng2021Transporter}.
Multi-camera systems extend perception to room-scale
human activity, but have primarily targeted offline
reconstruction or dataset creation rather than
live robot control~\cite{joo2016Panoptic, kim2025ParaHomea, grauman2024EgoExo4D}.
Using these systems for robot interaction requires
both accurate tracking under occlusion and low latency.
Multiple camera views help maintain tracking
when furniture or people obstruct individual views,
but also increase the amount of data to process.
This additional workload can delay the state estimates
available to robot controllers, limiting their ability
to respond promptly to human activity.
\begin{figure}[!t]
\centering
\includegraphics[width=\columnwidth]{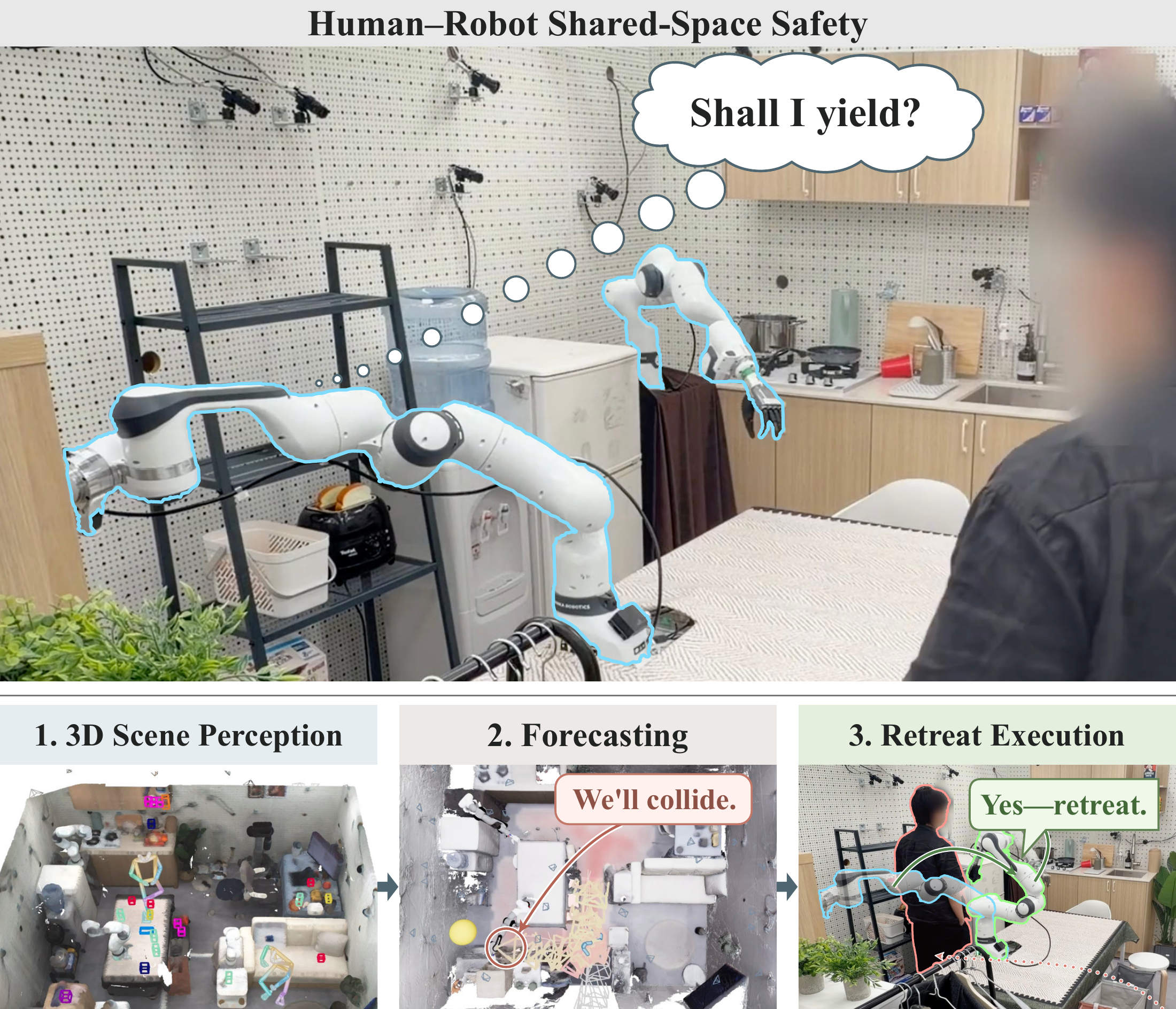}
\caption{
\textsc{OmniRobotHome} is a room-scale system for studying human-robot interaction in a furnished home. Its 48 hardware-synchronized RGB cameras provide real-time, markerless, occlusion-robust 3D perception of people and objects for three manipulators, with all perception and robot states expressed in a shared world frame. Using our system, we record human activity in the room to build a behavior memory for motion forecasting. The figure illustrates forecast-based yielding as one example of HRI supported by the system: live human tracking and motion forecasts inform the robot's decision to retreat before disrupting human activity.}

\label{fig:teaser}
\end{figure}

To address these challenges, we present \textsc{OmniRobotHome}, a room-scale testbed that continuously streams human and object state to robot controllers. A synchronized 48-camera system and distributed processing pipeline provide persistent, real-time 3D tracking throughout a furnished home, while dedicated workspace views provide the object-level 6D poses needed for manipulation. These room- and workspace-scale estimates are unified in a common world frame shared with the manipulators. The system also collects and learns from recorded human activity,
stored as room-specific behavior memory, to predict future human motion. These predictions enable robots to yield proactively
before interfering with human activity.

We demonstrate two interactions in \textsc{OmniRobotHome}:
assistive manipulation guided by nonverbal instructions
from across the room and proactive yielding to people
moving through the shared spaces.
In the assistance study, a robot interprets pointing
and gaze to select objects and destinations
for pick-and-place tasks.
In the yielding study, a manipulator uses human-motion
forecasts to yield before interfering with human activity.
Controlled experiments vary camera configurations,
perception latency, behavior-memory size,
forecasting methods, and ways of giving instructions.
The results show that timely observations and behavior-memory-based
forecasting reduce disruption, while coverage of human activity
areas improves nonverbal intent recognition.
These studies provide practical guidance for system design
and demonstrate the testbed's utility for evaluating
human–robot assistance in living environments.

Our contributions are threefold.
(1) We introduce OmniRobotHome, a furnished-home testbed
for studying human–robot assistance across the living space.
(2) We integrate synchronized multi-camera capture,
distributed low-latency perception, and human-motion
forecasting learned from room-specific behavior memory
to support robot interaction.
(3) We conduct controlled studies showing how
perception and forecasting design choices
affect proactive yielding and assistive manipulation.

\section{Related Work}
\label{sec:relworks}

\paragraph{Room-Scale Human and Object Sensing.}
Multi-view sensing helps maintain 3D estimates of people and objects when furniture or other people occlude individual views.
Panoptic Studio~\cite{joo2016Panoptic} uses synchronized multi-view capture to reconstruct human motion during social interaction.
Related capture systems recover human and object motion during grasping~\cite{taheri2020GRAB}, everyday home activities~\cite{kim2025ParaHomea}, and interactions involving multiple people and objects~\cite{zhang2024hoi}.
These systems emphasize offline reconstruction and dataset creation rather than live robot control.
Online systems such as OpenPTrack~\cite{munaro2016OpenPTrack} and Hi-ROS~\cite{guidolin2023HiROS} provide real-time multi-person tracking through RGB-D camera networks.
For objects, FoundationPose~\cite{wen2024FoundationPose} estimates and tracks 6D poses from RGB-D observations.
While these works focus on reconstructing human–object interactions or tracking people or objects, OMNIROBOTHOME integrates these capabilities to support live robot interaction across a furnished living space. 
It combines room-scale human pose estimation and coarse object tracking with workspace-level 6D object pose tracking, delivering real-time state estimates to robot controllers in a shared coordinate frame.

\paragraph{Robotic Manipulation and Assistance Systems.}
Platforms for robot learning have scaled in task diversity and data volume.
ALOHA and Mobile ALOHA support bimanual skill learning through teleoperation~\cite{zhao2023Learning, fu2024Mobile}.
Manipulation datasets broaden training experience across real environments~\cite{khazatsky2025DROID} and robot embodiments~\cite{collaboration2025Open}.
Language-guided systems connect user instructions to executable robot actions~\cite{ahn2022Can, wang2025MOSAIC}.
Where these systems emphasize skill learning and task execution, \textsc{OmniRobotHome} provides a testbed with persistent room-scale perception for developing HRI and conducting controlled studies of how perception affects interaction outcomes.

\paragraph{Human-Robot Interaction and Motion Prediction.}
Robot assistance in shared spaces involves interpreting human intent and avoiding interference with ongoing activity.
Human observations guide robot assistance through gaze-conditioned manipulation~\cite{zuo2026Gaze2Act} and skeleton-based intent recognition for collaborative pushing~\cite{solak2025Contextaware}.
For motion adaptation, reactive controllers respond to sensed obstacles~\cite{bhardwaj2021STORM}, and multi-camera skeleton tracking supports collision avoidance through control barrier functions~\cite{ferraguti2020MulticameraSafety}.

Anticipatory approaches account for future human behavior when selecting robot actions.
Human-aware planning in sensor-equipped homes schedules robot activity around expected human actions~\cite{cirillo2012AmbientPlanning}.
Learned motion models predict pedestrian trajectories~\cite{salzmann2020trajectron_simplified, bae2024singulartrajectory_simplifed} and future body poses from motion history~\cite{mao2020History}.
For collaborative manipulation, ManiCast uses learned motion forecasts in model predictive control~\cite{manicast2023}, and InteRACT conditions human intent prediction on robot actions~\cite{kedia2024InteRACT}.
While these works advance intent interpretation and motion prediction, \textsc{OmniRobotHome} integrates persistent room-scale perception with human-motion forecasts learned from room-specific behavior memory to support anticipatory and assistive interaction across the living space.

\newcommand{\systemoverviewcontent}{%
  \centering
  \includegraphics[width=\linewidth]{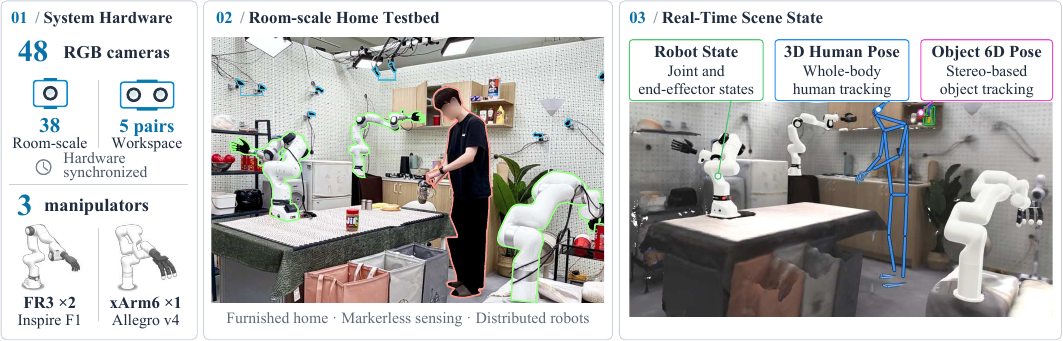}
  \vspace{-15pt}
  \captionof{figure}{%
    \textbf{System overview of \textsc{OmniRobotHome}.}
    \textbf{(1)} A hardware-synchronized RGB array comprises 38 room-scale
    cameras and five workspace stereo pairs (48 cameras total), while three
    manipulators are distributed across the activity areas.
    \textbf{(2)} The 23.1\,m$^2$ furnished testbed supports natural everyday
    activity without body markers, object tags, or wearable devices.
    \textbf{(3)} Real-time perception and robot feedback provide 3D human pose,
    6D object pose, and robot states in a unified world frame for downstream
    interaction.
  }
  \label{fig:system_overview}
}

\ifdefined\methodpreviewfirstpagefigure
  \twocolumn[{%
    \begin{minipage}{\textwidth}
      \systemoverviewcontent
    \end{minipage}
    \vspace{12pt}
  }]
\else
  \begin{figure*}[t!]
    \systemoverviewcontent
  \end{figure*}
\fi

\section{\textsc{OmniRobotHome}}
\label{sec:system}

We design \textsc{OmniRobotHome} to enable and study assistive robotic manipulation across a living space. The system connects three layers. (1) \textbf{Hardware} observes and acts throughout the room using full-space camera coverage and manipulators distributed across work areas. Using this infrastructure, the system records human activity to build \textbf{behavior memory} of how people move and interact. (2) \textbf{Perception} transforms RGB observations into a real-time representation of activity throughout the room, estimating 3D human poses and 6D poses of manipulation targets. \textbf{Prediction} uses recent human motion and priors from behavior memory to forecast future motion. (3) \textbf{Interaction} turns perceived and predicted states into robot behavior, yielding to an approaching person or executing assistive manipulation tasks.

\subsection{Room-Scale Hardware}
\label{sec:hardware_configuration}

\textsc{OmniRobotHome} occupies a 23.1\,m$^2$ furnished space with functional zones for cooking, storage, and everyday living. The space contains standard household appliances, furniture, utensils, and decorative objects. People and objects carry no markers, tags, or wearable devices.

Severe occlusions in un-instrumented home environments make reliable room-scale observation challenging, necessitating dense multi-view coverage. Accordingly, 38 of the 48 hardware-synchronized RGB cameras are distributed around the room's perimeter to maintain overlapping global views. The remaining ten cameras form five stereo pairs dedicated to the manipulation workspaces. To handle high-throughput visual data in real time, the 48-camera room-scale array is distributed across 15 edge PCs (two to four cameras per node), each equipped with a single RTX 4070 SUPER or RTX 3090 GPU. Human perception is distributed across the room-scale camera nodes, while five of the edge PCs additionally run stereo-based object 6D pose estimation for task-relevant regions.

Scaling real-time perception to 48 synchronized cameras is itself a systems challenge. At $2048 \times 1536$ and 30\,Hz, the array generates approximately 13.6\,GB/s of image data, so we keep pixels at the edge: each node performs detection and 2D pose estimation locally and transmits only keypoints, reducing human-perception traffic to about 2\,MB/s. Reliable operation additionally requires dedicated jumbo-frame NICs with enlarged kernel receive buffers and link-scoped routes between the nodes and the host; otherwise, traffic is routed through the building network, where we measured 0.4--1\% packet loss and up to 8\,ms of jitter. Because such misconfigurations may remain silent at the application level, an automated provisioning check verifies these settings across all nodes.

Robotic assistance is provided by three heterogeneous manipulators distributed across the activity areas. The system supports multiple arm and hand embodiments through a shared perception and manipulation framework. Our setup includes Franka Research 3 and xArm6 manipulators equipped with Inspire F1 and Allegro-v4 dexterous hands. The manipulators are positioned at locations naturally used by a resident, with overlapping working volumes that enable inter-robot handovers and continuous assistance across functional zones.
For spatial calibration, camera extrinsics are recovered through multi-view bundle adjustment, while robot bases are integrated via hand-eye calibration. This yields a mean visual reprojection error of 0.37\,px and a base registration error of 1.9\,mm (0.3$^\circ$). To align the system temporally, an external hardware trigger synchronizes all camera exposures, and robot states are queried at the corresponding camera time\-stamps, keeping synchronization error below 16.7\,ms.

\subsection{Real-Time Scene Perception}
\label{sec:real_time_perception}
\ifdefined\methodreviewred\methodreviewred\fi
{\color{black}\noindent\textbf{3D Human Pose Estimation.} Our system estimates the 3D whole-body pose of all persons in real time at approximately 25\,FPS, which downstream prediction and interaction modules use. Edge nodes extract 2D keypoints from incoming camera streams via a two-stage pipeline. On each node, a TensorRT-optimized INT8 YOLO26~\cite{yolo26_ultralytics} detector first localizes human bounding boxes, which are passed to an FP16 RTMPose~\cite{jiang2023RTMPose} model to estimate $133$ COCO-WholeBody keypoints per person. The 2D coordinates and confidence scores are streamed to the central host. %
\par}
{\color{black}On the central host, incoming 2D detections are first matched against active 3D tracks, while unassigned keypoints dynamically seed new candidate tracks to accommodate newly appearing individuals. %
For each confirmed person, whole-body keypoints are triangulated using Huber-weighted iteratively reweighted least squares (IRLS) to reduce the influence of outlier views. %
Finally, a per-joint One Euro filter smooths the estimated 3D trajectories. %
\par}

\suppressfloats[t]

\begin{figure}[t]
\centering\color{black}
\input{figures/33_forecast_pair/grid}
\captionof{figure}{
\textbf{Human-motion forecasts.}
At successive timesteps, the model uses the latest $3$\,s history to sample $N{=}32$ plausible $5$\,s futures. Gray skeletons show the observed history, red shows the current motion, and pale skeletons show the sampled futures.}
\label{fig:human_forecast}
\end{figure}

\noindent\textbf{Coarse 3D Object Tracking.} The system tracks the approximate 3D state of each object throughout the room. For each object class, we construct a fixed visual prototype embedding (VPE) by averaging and normalizing visual-prompt embeddings from multiple annotated reference images. At each synchronized timestep, edge nodes use YOLOE~\cite{yoloe} with these VPEs to detect objects in each view. The central host associates same-class detections across views based on their confidence and epipolar consistency. For each validated cluster, the host triangulates the object center, reconstructs its occupied volume within a class-specific search region, and fits an oriented 3D cuboid. Each resulting track provides the object's class, approximate center, and extent. We use the estimated location to determine which robot workspace contains the object.

\noindent\textbf{6D Object Pose Tracking.} Once coarse 3D tracking locates an object in a robot workspace, the stereo pipeline estimates and tracks its precise 6D pose at approximately 16\,Hz. The pipeline requires a template mesh for each object. We use the object's CAD model when available and otherwise reconstruct the mesh offline with MV-SAM3D~\cite{li2026mv}. At runtime, a TensorRT-optimized Fast-FoundationStereo~\cite{wen2026fastfoundationstereo} model generates dense metric depth from rectified stereo views, while YOLOE segments the object. FoundationPose~\cite{wen2024FoundationPose} uses these outputs and the template mesh to initialize the object's 6D pose. After initialization, it switches to tracking mode and updates the pose across frames. During 6D tracking, periodic YOLOE masks validate the active track and reinitialize FoundationPose after tracking failures.

\subsection{Behavior Memory and Human Motion Forecasting}
\label{sec:behavior_memory}
\label{sec:human_motion_forecasting}
\label{sec:prediction}

\noindent\textbf{Behavior Memory.}
We use the same multi-camera infrastructure for online perception and
the collection of room-specific human activity, building a
\emph{behavior memory} for anticipatory robot assistance.
We record unscripted activities as 3D full-body pose trajectories,
capturing how people navigate and interact within the living space.
For forecasting, we represent the memory as history--future pairs
$\mathcal{M}=\{(h_i,x_{1,i})\}_{i=1}^{M}$, where $h_i$ is a $3$\,s
tracking history sampled at $15$\,Hz and $x_{1,i}$ is its observed
$5$\,s future.
This memory provides both training data for the forecaster and
candidate futures for motion recombination.

\noindent\textbf{Motion Forecaster.}
We extend room-scale perception to future human motion with
a conditional rectified-flow forecaster~\cite{liu2022rectifiedflow}
trained on the behavior memory $\mathcal{M}$.
Given a tracking history $h$, the model samples $N{=}32$ plausible futures.
Motion is represented by a 3D mid-hip root and 13 root-relative joints,
together with per-joint triangulation confidence derived from
multi-view reprojection quality.
Missing observations are replaced with learned null tokens, and
confidences are provided as additional input channels.
A transformer history encoder $E_\phi$ summarizes the history into
a conditioning vector $E_\phi(h)$.
Conditioned on $E_\phi(h)$ via AdaLN, a diffusion transformer
(DiT) decoder $f_\theta$~\cite{peebles2023scalable} learns a
rectified flow from Gaussian noise $z_0 \sim \mathcal{N}(0,I)$
to future motion $x_1$, with the root trajectory expressed
relative to the last observed mid-hip.

We encourage diverse predictions through best-of-$K$
training~\cite{rupprecht2017learning}.
For each training window, we draw $K{=}16$ noise seeds and update
only the best-matching seed using a confidence-weighted
velocity-matching loss:
\begin{equation}
\mathcal{L} =\!\!\!\! \mathop{\mathbb{E}}\limits_{t,(h,x_1)}
\min_{k\in[K]}\!
\Big\lVert\,w \odot
\Big(\!f_\theta\!\big(x_t^{(k)}\!\!,\!\,t,\,\!E_\phi(h)\big)\!
-\! \big(x_1 \!-\! z_0^{(k)}\big)\!\Big)\Big\rVert^{2},
\end{equation}
where $x_t^{(k)}=(1-t)z_0^{(k)}+t\,x_1$ for
$t\sim\mathcal{U}(0,1)$, and $w$ masks low-confidence joints
and weights the rest by triangulation confidence.
At inference, we reuse the encoded history across all $N$ samples
and integrate the flow with $S{=}8$ Euler steps.

Motion recombination expands training beyond the single future
observed after each history.
We retrieve futures from similar situations in
$\mathcal{M}$ using nearest-neighbor search and connect them to
the observed motion with a duration head and a bidirectional
rectified-flow in-betweener.
We reject implausible recombinations, such as foot
skating or unnatural heading reversals.
Farthest-point sampling over future endpoints selects diverse
continuations, broadening future-motion coverage while preserving
room-scale plausibility.

\subsection{Interaction}
\label{sec:interaction}
\label{sec:manipulation}
\begin{figure}[t]
  \centering
  \color{black}
  \includegraphics[width=\columnwidth]{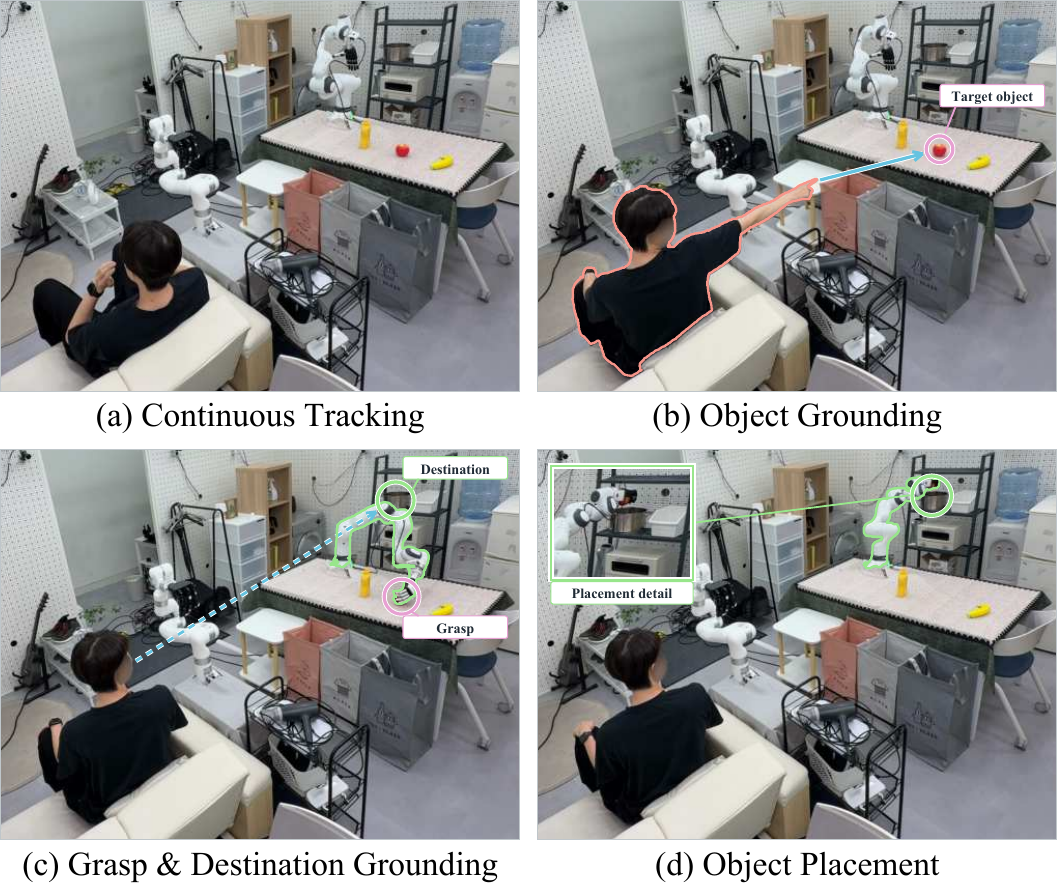}
  \captionof{figure}{
    \textbf{Spatial-cue-conditioned assistance.}
    (a) The system continuously tracks the resident.
    (b) A pointing cue selects the apple as the manipulation target.
    (c) While the robot grasps the apple using its live 6D pose,
    the user's gaze specifies the destination.
    (d) The robot transfers the apple and releases it at the selected
    destination.
  }
  \label{fig:demo_examples}
\end{figure}

We use the system's perception capabilities to support
anticipatory yielding, robot manipulation, and nonverbal
spatial commands.

\noindent\textbf{Non-Disruptive Yielding.}
We use human-motion forecasts to anticipate interference and
trigger robot yielding in shared workspaces.
For each arm, the controller evaluates the closest person's
$N{=}32$ predicted trajectories over a $1.5$\,s horizon.
It computes the minimum distance between per-link robot
collision spheres and human body capsules across all predicted
futures.
When this distance falls below $0.2$\,m, the arm moves to a
predefined non-interfering posture and waits.
It resumes once the distance stays above $0.3$\,m for $1$\,s,
avoiding rapid switching between yielding and execution.

\noindent\textbf{Robot Manipulation.}
We enable object manipulation by combining tracked 6D object
poses with an offline Dexonomy~\cite{chen2025dexonomy} grasp
library.
Given a target object, we transform candidate grasps into the
robot base frame, filter them for geometric and kinematic
feasibility, including collision clearance, and use
cuRobo~\cite{sundaralingam2023curobo} to plan the arm motion.
For transport and placement, we plan motion between
manually specified waypoints.

\noindent\textbf{Nonverbal Intent Recognition.}
We interpret nonverbal instructions by identifying spatial
targets indicated through gaze or pointing.
Using the tracked 3D human pose, we estimate gaze rays from
head and torso geometry and pointing rays from upper-body
and index-fingertip keypoints.
We calibrate the gaze parameters by having participants
look at known targets at various locations.
Once a cue stabilizes, we select the target with the smallest angular deviation from the ray and confirm the selection once it has remained consistent for $0.5$\,s.

{\experimentreviewred
\begin{figure}[t]
  \centering\color{black}
  \includegraphics[width=\columnwidth]{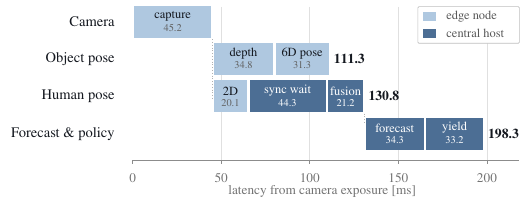}
  \captionof{figure}{\textbf{Pipeline latency relative to camera exposure.}
  Stage runtimes measured with 38 cameras tracking one person. At 1.4\,m/s a
  person travels 28\,cm during the latency to an interaction decision.}
  \label{fig:latency_paths}
\end{figure}

\makeatletter\global\@topnum\z@\makeatother
\begin{table}[t]
  {\color{black}\noindent
\begin{minipage}[t]{\columnwidth}
\vspace{0pt}
\centering
\scriptsize
\setlength{\tabcolsep}{0pt}
\renewcommand{\arraystretch}{1.2}

\captionof{table}{\textbf{Perception accuracy and runtime comparison.}}
\label{tab:percep_ablation}

\begin{tabular*}{\linewidth}{
  @{\extracolsep{\fill}}
  lcr
  @{\hspace{2pt}}|@{\hspace{2pt}}
  lcr
  @{\hspace{2pt}}|@{\hspace{2pt}}
  lcr
  @{}
}
\toprule
Triangulation & Err [px]$\downarrow$ & [ms]
  & Depth & IoU$\uparrow$ & [ms]
  & 6D Pose & IoU$\uparrow$ & [ms] \\
\specialrule{\lightrulewidth}{1.4pt}{1.4pt}
DLT          & 16.8 & \textbf{5.1}
  & MoGe~\cite{wang2025moge2accuratemonoculargeometry} & 0.12 & 126
  & GoTrack~\cite{nguyen2025gotrack} & 0.11 & 1268 \\
+ RANSAC     & 14.2 & 24.0
  & DA3~\cite{lin2025depthanything3} & 0.12 & 224
  & MegaPose-T~\cite{labbe2022megapose} & 0.37 & 273 \\
             &      &
  & FS~\cite{wen2025FoundationStereo} & \textbf{0.73} & 579
  & SAM-6D~\cite{lin2024sam6d} & 0.65 & 256 \\
\specialrule{\lightrulewidth}{1.4pt}{1.4pt}
\textbf{Ours} & \textbf{13.9} & 10.1
  & \textbf{Ours} & 0.71 & \textbf{34.8}
  & \textbf{Ours} & \textbf{0.71} & \textbf{31.3} \\
\bottomrule
\end{tabular*}
\end{minipage}

\floattrim{-4pt}
}
\end{table}
\begin{table}[t]
  \color{black}
        \centering
    \renewcommand{\arraystretch}{1.2}
    \scriptsize
    \setlength{\tabcolsep}{2pt}
    
   \captionof{table}{\textbf{Impact of perception latency and forecasting on non-disruptive yielding.}}
    \label{tab:safety_main}
    \begin{tabular*}{\linewidth}[h]{@{\extracolsep{\fill}}l ccc c@{}}
      \toprule
      & $+$1.0\,s & $+$0.5\,s & $+$0.0\,s
        & \textbf{\makecell{$-$1.5\,s (Ours, forecast)}} \\
      \midrule
      Disrupt\,\%~$\downarrow$ & 31.65 & 29.50 & 26.29 & \textbf{7.32} \\
      Retreat\,\%~$\downarrow$ & 15.18 & 15.36 & 15.49 & 17.30 \\
      \bottomrule
    \end{tabular*}
  \floattrim{-10pt}  %
\end{table}

\section{EXPERIMENTS}
\label{sec:experiments}

\subsection{Perception Performance and System Latency}
\label{sec:exp_perception_and_yielding}
\label{sec:realtime}
To validate the efficiency of our system design, we profile the latency of each perception path, measured from camera exposure to the interaction-policy decision, excluding subsequent robot motion (\figref{fig:latency_paths}).
Captured images reach the shared-memory buffer in 45.2\,ms, after which the human and object branches proceed in parallel. 
On the object branch, stereo depth estimation takes 34.8\,ms, followed by 31.3\,ms for 6D pose estimation, yielding an object pose at 111.3\,ms after camera exposure.
Segmentation takes 14.7\,ms and runs periodically for track initialization and validation, so it does not lie on the per-frame critical path.

On the human branch, per-camera 2D whole-body perception (20.1\,ms), the wait
for sufficient views (44.3\,ms), and multi-view fusion (21.2\,ms) produce a
fused 3D human pose at 130.8\,ms. This wait is dominated by the slowest of the
22 views required to promote a frame rather than by network transfer, which
accounts for less than 1\,ms. Human-motion forecasting then takes
34.3\,ms, making predicted futures available at 165.1\,ms. The yielding test
takes an additional 33.2\,ms and issues a retreat decision at 198.3\,ms.
The human branch therefore determines the interaction-policy latency.
Gaze and pointing incur little additional computation once the 3D human pose
is available.

We evaluated alternative perception modules across key stages to develop an efficient perception stack for real-time robot interaction (\tabref{tab:percep_ablation}). 
For triangulation under outlier views, we compared DLT, RANSAC, and Huber-reweighted IRLS using held-out-view reprojection error.
Depth and pose estimators were integrated into the 6D object-pose pipeline and evaluated by IoU between the segmented mask and the mesh rendered from the estimated pose.
These comparisons guided the choice of components for the online pipeline based on both accuracy and runtime.

We use \textsc{OmniRobotHome} to study how perception latency
and anticipation affect non-disruptive yielding
(\tabref{tab:safety_main}).
We evaluate 130 held-out human-motion sessions from four
participants, each lasting 20--30\,s and following a household
scenario with specified start and end points and intermediate
waypoints.
For controlled comparisons, we replay each session against
a fixed robot trajectory at 40 temporal offsets.
After filtering out cases with insufficient human--robot
overlap, 4,976 evaluation cases remain.

We compare three reactive settings that use only the observed
human pose under additional delays of 0, 0.5, or 1.0\,s
with forecast-based yielding using a 1.5\,s prediction horizon.
The reactive delay settings use a 0.5\,m trigger margin,
while forecast-based yielding uses the 0.2\,m margin
described in \secref{sec:interaction}.
We also evaluate reactive yielding at 0.2\,m for a
matched-margin comparison.
We define \emph{Disrupt} as the fraction of evaluation cases
in which the moving robot enters the person's recorded path,
and \emph{Retreat} as the fraction of execution time spent
in the yielded state.

Forecasting substantially reduces \emph{Disrupt} from 31.6\%
to 7.32\% compared with reactive yielding at the same
0.2\,m trigger margin, with \emph{Retreat} of 17.30\%.
For reactive yielding with a 0.5\,m margin, increasing the
additional delay from 0 to 1.0\,s raises \emph{Disrupt}
from 26.29\% to 31.65\%, while \emph{Retreat} remains
nearly unchanged.
These results demonstrate the value of low-latency perception
and anticipation for reducing interference with human activity.

\subsection{Spatial Coverage under Occlusion}
\label{sec:density}

Leveraging its dense camera array, \textsc{OmniRobotHome} allows us to examine how occlusion changes the camera density required for reliable human-pose reconstruction in a furnished room (\figref{fig:density_views}b). 
We define \emph{reconstructability} as the fraction of joints observed from at least two views. Fewer than 10 cameras make nearly all joints reconstructable under idealized geometric visibility, while the furnished setting requires 19 cameras to reach 95\% reconstructability for both body and feet. Occlusion also reduces view redundancy unevenly across joint groups (\figref{fig:density_views}d). With 38 cameras, 84.0\% of foot joints are observed from at least eight views, compared with 90.1\% of body joints and 99.0\% of hand joints. Lower-body joints receive the least multi-view support in the furnished environment.

Additional views continue to improve localization after two-view
reconstructability saturates (\figref{fig:density_views}c). Under an assumed pixel noise of $\sigma_{\mathrm{px}}{=}1.0$ px, median analytic 3D uncertainty decreases with camera density, reaching 0.98, 1.31, and 0.81\,mm for body, feet, and hands at 38 cameras. Thus, additional views improve localization precision even when most joints are already reconstructable.

We next examine whether spatial coverage affects downstream robot yielding (\figref{fig:density_views}e). We define \emph{spatial coverage} as the fraction of the room observed by at least three cameras after accounting for occlusion. Across 72 configurations with 4--38 cameras, \emph{Disrupt} decreases from roughly 40\% to 10\% as coverage increases, while \emph{Retreat} increases from roughly 5\% to 17\%. Greater spatial coverage therefore supports less disruptive yielding at the cost of more time spent yielded.

\subsection{Behavior Memory and Forecasting}
\label{sec:safety}
\begin{figure}[t]
  \centering
  \includegraphics[width=\columnwidth]{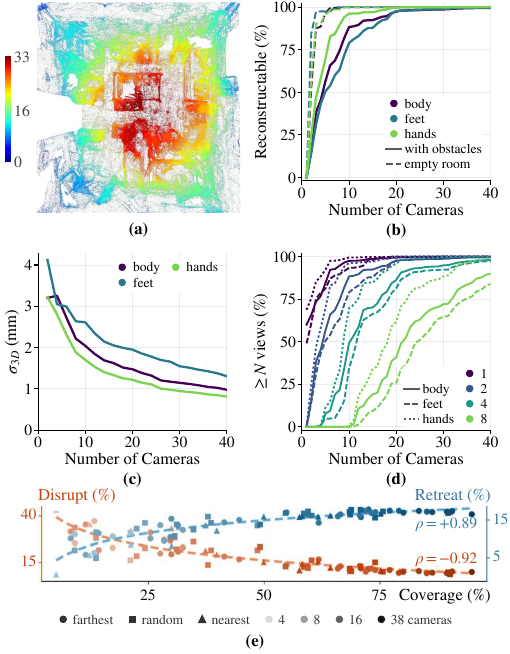}\\[0.5mm]
  \vspace{-2mm}
  {\color{black}
  \caption{\textbf{Spatial coverage and joint reconstructability
  across camera densities.}
  \textbf{(a)} Per-voxel observation count across the room for 38 cameras.
  \textbf{(b)} Joint reconstructability across occlusion conditions.
  \textbf{(c)} Analytic 3D localization uncertainty
  ($\sigma_{\mathrm{3D}}$).
  \textbf{(d)} Joints with at least $N$ detected views.
  \textbf{(e)} Correlation between room coverage and non-disruptive
  yielding.}
  \label{fig:density_views}
  }
\end{figure}

We use \textsc{OmniRobotHome} to study how accumulated human
activity and forecasting design improve motion prediction
and proactive yielding.
The full behavior memory contains 4\,h of unscripted activity
from 10 participants.
We evaluate forecasting on the same 130 held-out sessions used
for yielding, collected from four participants not included
in the behavior memory.

We measure forecasting accuracy with minADE, the minimum
average displacement error among the $N$ sampled futures
over the prediction horizon and 13 joints.
Root-minADE computes the same metric using only the mid-hip root.
APD measures sample diversity as the mean pairwise distance
between terminal root positions.
Recall@0.3 measures future coverage as the fraction of windows
with at least one sample within 0.3\,m of the ground truth
at every predicted frame.

\noindent\textbf{Behavior Memory and Training.}
Larger behavior memory improves forecasting accuracy,
diversity, and future coverage
(\tabref{tab:data-learning-ablation}).
These improvements substantially reduce \emph{Disrupt}
with only a slight increase in \emph{Retreat}.
Best-of-$K$ training improves sample diversity as well as
minADE and Recall@0.3 compared with training without it.
Motion recombination further improves all three forecasting
metrics and reduces \emph{Disrupt}, demonstrating the benefit
of expanding training with alternative futures drawn from
the same behavior memory.

\noindent\textbf{Forecaster Comparison.}
Our forecaster achieves the lowest root-minADE in the
comparison with Motion Matching and forecasting baselines
(\tabref{tab:forecast_safety}).
The public Trajectron++ checkpoint achieves lower
\emph{Disrupt} but has higher prediction error and
\emph{Retreat}.
Among models trained on the same room-specific data,
our forecaster reduces both \emph{Disrupt} and \emph{Retreat}
relative to the retrained baselines.
Qualitatively, our forecasts better respect the room layout
than those of the baselines, with fewer trajectories passing
through tables or extending beyond room boundaries.
This comparison shows that our forecasting design improves
motion prediction while supporting less disruptive interaction
with less time spent yielded.

\begin{table}[t]
  \color{black}
  \centering
\renewcommand{\arraystretch}{1.2}
\scriptsize
\setlength{\tabcolsep}{2pt}
\captionof{table}{\textbf{Behavior memory and forecaster ablations.}}
\label{tab:data-learning-ablation}
\begin{tabular*}{\linewidth}[t]{l !{\hspace{0.5em}\vrule} @{\extracolsep{\fill}} ccc !{\vrule} cc}
    \toprule
    Settings & \makecell{minADE$\downarrow$} & \makecell{APD$\uparrow$}
      & \makecell{Recall@0.3$\uparrow$}
      & \makecell{Disrupt\%~$\downarrow$} & \makecell{Retreat\%~$\downarrow$} \\
    \midrule
    1\,Hr           & 0.49 & 0.32 & 0.39 & 26.57 & \textbf{12.69} \\
    2\,Hrs          & 0.44 & 0.47 & 0.45 & 18.55 & 15.00 \\
    4\,Hrs          & 0.37 & 0.72 & 0.53 & 13.97 & 16.11 \\
    \midrule
    4\,Hrs (w/o best-of-$K$) & 0.48 & 0.27 & 0.41 & 16.82 & 14.88 \\
    \midrule
    4\,Hrs + Aug (Ours)     & \textbf{0.28} & \textbf{1.12} & \textbf{0.69}
      & \textbf{7.32} & 17.30 \\
    \bottomrule
\end{tabular*}

  \floattrim{-4pt}
\end{table}
\begin{table}[t]
  \color{black}
  \centering
  \renewcommand{\arraystretch}{1.2}
  \scriptsize
  \setlength{\tabcolsep}{2pt}
  \captionof{table}{\textbf{Comparison with motion-forecasting baselines (95\,\% CI).}}
  \label{tab:forecast_safety}
  \begin{tabular*}{\linewidth}[h]{l !{\hspace{0.5em}\vrule\hspace{0.2em}} @{\extracolsep{\fill}} ccc}
    \toprule
    Forecaster & root-minADE~$\downarrow$ & Disrupt\,\%~$\downarrow$ & Retreat\,\%~$\downarrow$ \\
    \midrule
    Motion Matching            & 0.35 \tiny(0.32,\,0.38) & 17.58 \tiny(13.34,\,22.24) & 19.38 \tiny(17.71,\,21.08) \\
    Trajectron++~\cite{salzmann2020trajectron_simplified}  & 0.40 \tiny(0.37,\,0.43) & \textbf{5.49} \tiny(2.65,\,8.88) & 27.90 \tiny(25.49,\,30.37) \\
    \hspace{1em} retrained     & 0.27 \tiny(0.25,\,0.29) & 15.13 \tiny(10.71,\,19.78) & 24.20 \tiny(22.35,\,26.08) \\
    SingularTraj.~\cite{bae2024singulartrajectory_simplifed}  & 0.33 \tiny(0.31,\,0.36) & 15.86 \tiny(11.32,\,20.73) & 25.74 \tiny(23.24,\,28.24) \\
    \hspace{1em} retrained     & 0.22 \tiny(0.21,\,0.24) & 16.46 \tiny(11.80,\,21.02) & 23.78 \tiny(21.62,\,26.06) \\
    \midrule
    \textbf{Ours} & \textbf{0.20} \tiny(0.19,\,0.22) & 7.32 \tiny(4.91,\,9.97) & \textbf{17.30} \tiny(15.59,\,19.03) \\
    \bottomrule
  \end{tabular*}
\end{table}

\begin{table*}[t]
  {\color{black}\footnotesize
\centering
\caption{\textbf{Cue interpretation across camera configurations and cue types.}}
\label{tab:cue_camera_ablation}
\begin{tabularx}{\textwidth}{
  c
  >{\raggedright\arraybackslash}X
  c c
  c c
  c c
  c c
  c c
}
  \toprule
    & & & &
    \multicolumn{2}{c}{No commit / 50} &
    \multicolumn{2}{c}{Correct commits / 50} &
    \multicolumn{2}{c}{Error ($^\circ$)} &
    \multicolumn{2}{c}{Median commit time (s)} \\
  \cmidrule(lr){5-6}\cmidrule(lr){7-8}\cmidrule(lr){9-10}\cmidrule(lr){11-12}
    Cameras &
    \multicolumn{1}{c}{Region / selection} &
    \makecell{3-view\\coverage} &
    \makecell{Two-stage\\/ 50}
    & Gaze & Pointing & Gaze & Pointing & Gaze & Pointing & Gaze & Pointing \\
  \midrule
    38 & Full room       & 0.83 & 37 &  3 &  0 & 36 & 49 &  7.1 & 2.4 & 0.77 & 0.83 \\
    16 & Full room (FPS) & 0.81 & 35 &  4 &  1 & 34 & 49 & 10.2 & 2.6 & 0.78 & 0.83 \\
    8 & Full room (FPS) & 0.71 & 34 &  3 &  2 & 34 & 47 &  9.6 & 2.7 & 0.80 & 0.83 \\
    4 & Full room (FPS) & 0.41 & 20 & 11 & 14 & 24 & 33 & 11.6 & 3.0 & 0.90 & 0.88 \\
    \midrule
    8 & Lounge region & 0.67 & 32 &  9 &  3 & 31 & 46 & 10.1 & 2.5 & 0.80 & 0.83 \\
    4 & Lounge region & 0.42 & 18 & 18 & 18 & 20 & 29 & 10.6 & 3.1 & 0.92 & 0.87 \\
    \midrule
    8 & Robot workspace & 0.34 & 5 & 45 & 43 & 5 & 6 & 16.8 & 4.1 & 0.33 & 0.87 \\
    4 & Robot workspace & 0.15 & 4 & 42 & 45 & 7 & 5 & 20.2 & 4.1 & 0.52 & 0.80 \\
  \bottomrule
\end{tabularx}
\floattrim{-10pt}
}
  \floattrim{-4pt}
\end{table*}

\subsection{Room-Scale Assistive Manipulation}
\label{sec:application}

We combine nonverbal intent recognition with robot manipulation
to demonstrate room-scale pick-and-place assistance.
Participants select an object and a placement destination
through gaze or pointing from freely chosen locations in the room.
Five participants perform ten trials each.
An LLM-generated protocol assigns one of three objects and
a cue type for the first stage, then one of three destinations
and a cue type for the second, balancing objects, destinations,
and cues across the 50 trials.

\noindent\textbf{Nonverbal Intent Recognition.}
With the full 38-camera human-perception setup, the system
correctly identifies both the requested object and destination
in 37/50 trials (\tabref{tab:cue_camera_ablation}).
Pointing produces more correct commits than gaze
(49/50 vs. 36/50) and a lower median angular error
($2.4^\circ$ vs. $7.1^\circ$).

\noindent\textbf{Physical Execution.}
We use these nonverbal requests to drive the robot end to end
without manual correction (\tabref{tab:manipulation_eval}).
Of the 37 trials with both targets correctly identified,
33 complete grasping and 27 complete placement, yielding
27/50 end-to-end successes.
Several transport failures involve unstable top-down grasps
on the mustard bottle's cap.
Median grasp and placement planning times are 4.8 and 0.1\,s,
respectively, with placement planning running concurrently
with grasp execution.
Median time from the go signal to robot return is 35.1\,s.

\noindent\textbf{Camera Count and Placement.}
We use the testbed's multi-view recordings to compare camera
configurations under the same nonverbal requests.
We reconstruct 3D poses using different camera subsets,
selecting spatially distributed subsets by farthest-point
sampling.
To study placement, we compare a lounge-focused layout with
a robot-workspace layout centered on the manipulation table.
Region-focused views are selected using scene-mesh visibility
and angular diversity around each region center.

Eight spatially distributed cameras preserve much of the
full setup's two-stage cue success (34/50 vs. 37/50),
despite reducing three-view coverage from 0.83 to 0.71.
Pointing error remains nearly unchanged at
$2.4^\circ$--$2.7^\circ$, while gaze error increases from
$7.1^\circ$ to $9.6^\circ$.
With four cameras, coverage falls to 0.41 and two-stage cue
success drops to 20/50.
Missing commits increase from 3/100 in the full setup to
25/100, although median pointing error among valid cues
remains low at $3.0^\circ$.
This contrast shows that sufficient multi-view support is
needed to keep cues observable, beyond improving the accuracy
of valid estimates.

At the same camera count, covering the regions where people
issue commands substantially improves intent recognition.
With eight cameras, the lounge-focused layout achieves
32/50 two-stage cue successes, compared with 5/50 for the
robot-workspace layout, where 88/100 cues produce no commit.
Reducing the lounge-focused layout to four cameras lowers
two-stage cue success to 18/50.
These results show that room-scale nonverbal assistance
requires overlapping views of human activity areas,
rather than concentrating cameras around the robot's
manipulation workspace.

\begin{table}[t]
  \color{black}
  \scriptsize
\centering
\renewcommand{\arraystretch}{1.2}
\setlength{\tabcolsep}{2pt}
\captionof{table}{\textbf{Physical manipulation performance.}}
\label{tab:manipulation_eval}
\begin{tabular*}{\columnwidth}[t!]{
  @{\extracolsep{\fill}}
  l r
  @{\hspace{3pt}}|@{\hspace{3pt}}
  l r
  @{\hspace{3pt}}|@{\hspace{3pt}}
  l r
  @{}
}
  \toprule
  \multicolumn{2}{c}{\textit{Task success}} &
  \multicolumn{2}{c}{\textit{Physical execution}} &
  \multicolumn{2}{c}{\textit{Timing (s)}} \\
  \cmidrule(r){1-2}
  \cmidrule(lr){3-4}
  \cmidrule(l){5-6}

  Stage & Success &
  Stage & Success &
  Component & Median \\
  \midrule

  Request correct   & 37/50 &
  Grasp             & 43/49 &
  Grasp plan        & 4.8 \\

  Grasp given req.  & 33/37 &
  Place             & 33/42 &
  Grasp exec.       & 8.4 \\

  Place given grasp & 27/33 &
  Grasp + place     & 33/49 &
  Place plan        & 0.1 \\

  End-to-end        & 27/50 &
                    &       &
  Place exec.       & 18.3 \\

  \bottomrule
\end{tabular*}
\floattrim{-10pt}

  \floattrim{-4pt}
\end{table}

\par}
{\color{black}
\section{Discussion}
\label{sec:conclusion}

\textsc{OmniRobotHome} provides a system-level testbed for room-scale HRI
in a furnished living environment.
Maintaining coverage despite occlusions from people and furniture
requires dense multi-view sensing, but the resulting image throughput
makes low-latency state delivery challenging.
Our system combines hardware-synchronized capture across 48 cameras
with distributed perception to process these observations in real time
and deliver persistent human and object state to robot controllers
in a shared world frame.
Using the recorded human activity, we also build a behavior memory
for room-specific motion forecasting.
This integration makes room-scale perception and forecasts available
within the robot control loop, enabling interaction beyond individual
manipulation workspaces.

We demonstrate this capability through studies of non-disruptive
yielding and nonverbal intent recognition for assistive manipulation.
With robot skills held fixed, we examine camera count and placement,
spatial coverage, state latency, behavior-memory scale,
forecasting methods, and interaction cues.
Low-latency observations and behavior-memory-based forecasting
reduce disruption during yielding.
For nonverbal assistance, covering the regions where people move
and issue cues improves intent recognition, while pointing
provides more reliable target selection than gaze.
Together, these findings provide practical guidance for system design
and demonstrate the value of \textsc{OmniRobotHome} as a testbed
for HRI in everyday living spaces.

\bibliographystyle{IEEEtran}
\bibliography{main,references}

\begin{thebibliography}{10}
\providecommand{\url}[1]{#1}
\csname url@rmstyle\endcsname
\providecommand{\newblock}{\relax}
\providecommand{\bibinfo}[2]{#2}
\providecommand\BIBentrySTDinterwordspacing{\spaceskip=0pt\relax}
\providecommand\BIBentryALTinterwordstretchfactor{4}
\providecommand\BIBentryALTinterwordspacing{\spaceskip=\fontdimen2\font plus
\BIBentryALTinterwordstretchfactor\fontdimen3\font minus
  \fontdimen4\font\relax}
\providecommand\BIBforeignlanguage[2]{{%
\expandafter\ifx\csname l@#1\endcsname\relax
\typeout{** WARNING: IEEEtran.bst: No hyphenation pattern has been}%
\typeout{** loaded for the language `#1'. Using the pattern for}%
\typeout{** the default language instead.}%
\else
\language=\csname l@#1\endcsname
\fi
#2}}

\bibitem{brohan2023RT1}
A.~Brohan, N.~Brown, J.~Carbajal, Y.~Chebotar, J.~Dabis, C.~Finn,
  K.~Gopalakrishnan, K.~Hausman, A.~Herzog, J.~Hsu, \emph{et~al.}, ``Rt-1:
  Robotics transformer for real-world control at scale,'' in \emph{RSS}, 2023.

\bibitem{kalashnikov2018QTOpt}
D.~Kalashnikov, A.~Irpan, P.~Pastor, J.~Ibarz, A.~Herzog, E.~Jang, D.~Quillen,
  E.~Holly, M.~Kalakrishnan, V.~Vanhoucke, \emph{et~al.}, ``Scalable deep
  reinforcement learning for vision-based robotic manipulation,'' in
  \emph{CoRL}, 2018.

\bibitem{shridhar2022CLIPort}
M.~Shridhar, L.~Manuelli, and D.~Fox, ``Cliport: What and where pathways for
  robotic manipulation,'' in \emph{CoRL}, 2021.

\bibitem{zeng2021Transporter}
A.~Zeng, P.~Florence, J.~Tompson, S.~Welker, J.~Chien, M.~Attarian,
  T.~Armstrong, I.~Krasin, D.~Duong, V.~Sindhwani, \emph{et~al.}, ``Transporter
  networks: Rearranging the visual world for robotic manipulation,'' in
  \emph{CoRL}, 2020.

\bibitem{joo2016Panoptic}
H.~Joo, T.~Simon, X.~Li, H.~Liu, L.~Tan, L.~Gui, S.~Banerjee, T.~Godisart,
  B.~Nabbe, I.~Matthews, T.~Kanade, S.~Nobuhara, and Y.~Sheikh, ``Panoptic
  {{Studio}}: {{A Massively Multiview System}} for {{Social Interaction
  Capture}},'' \emph{IEEE Trans. Pattern Anal. Mach. Intell.}, vol.~41, no.~1,
  pp. 190--204, 2019.

\bibitem{kim2025ParaHomea}
J.~Kim, J.~Kim, J.~Na, and H.~Joo, ``{{ParaHome}}: {{Parameterizing Everyday
  Home Activities Towards 3D Generative Modeling}} of {{Human-Object
  Interactions}},'' in \emph{CVPR}, 2025.

\bibitem{grauman2024EgoExo4D}
K.~Grauman, A.~Westbury, L.~Torresani, K.~Kitani, J.~Malik, T.~Afouras,
  K.~Ashutosh, V.~Baiyya, S.~Bansal, B.~Boote, \emph{et~al.}, ``Ego-exo4d:
  Understanding skilled human activity from first-and third-person
  perspectives,'' in \emph{CVPR}, 2024.

\bibitem{taheri2020GRAB}
O.~Taheri, N.~Ghorbani, M.~J. Black, and D.~Tzionas, ``{{GRAB}}: {{A Dataset}}
  of {{Whole-Body Human Grasping}} of {{Objects}},'' in \emph{ECCV}, 2020.

\bibitem{zhang2024hoi}
J.~Zhang, J.~Zhang, Z.~Song, Z.~Shi, C.~Zhao, Y.~Shi, J.~Yu, L.~Xu, and
  J.~Wang, ``Hoi-m3: Capture multiple humans and objects interaction within
  contextual environment,'' in \emph{CVPR}, 2024.

\bibitem{munaro2016OpenPTrack}
M.~Munaro, F.~Basso, and E.~Menegatti, ``Openptrack: Open source multi-camera
  calibration and people tracking for rgb-d camera networks,'' \emph{Robotics
  and Autonomous Systems}, vol.~75, pp. 525--538, 2016.

\bibitem{guidolin2023HiROS}
M.~Guidolin, L.~Tagliapietra, E.~Menegatti, and M.~Reggiani, ``Hi-ros:
  Open-source multi-camera sensor fusion for real-time people tracking,''
  \emph{Computer Vision and Image Understanding}, vol. 232, p. 103694, 2023.

\bibitem{wen2024FoundationPose}
B.~Wen, W.~Yang, J.~Kautz, and S.~Birchfield, ``Foundationpose: Unified 6d pose
  estimation and tracking of novel objects,'' in \emph{CVPR}, 2024.

\bibitem{zhao2023Learning}
T.~Z. Zhao, V.~Kumar, S.~Levine, and C.~Finn, ``Learning {{Fine-Grained
  Bimanual Manipulation}} with {{Low-Cost Hardware}},'' in \emph{RSS}, 2023.

\bibitem{fu2024Mobile}
Z.~Fu, T.~Z. Zhao, and C.~Finn, ``Mobile {{ALOHA}}: {{Learning Bimanual Mobile
  Manipulation}} with {{Low-Cost Whole-Body Teleoperation}},'' in \emph{CoRL},
  2024.

\bibitem{khazatsky2025DROID}
A.~Khazatsky \emph{et~al.}, ``{{DROID}}: {{A Large-Scale In-The-Wild Robot
  Manipulation Dataset}},'' in \emph{RSS}, 2024.

\bibitem{collaboration2025Open}
A.~O'Neill \emph{et~al.}, ``Open {{X-Embodiment}}: {{Robotic Learning
  Datasets}} and {{RT-X Models}},'' in \emph{ICRA}, 2024.

\bibitem{ahn2022Can}
M.~Ahn, A.~Brohan, N.~Brown, Y.~Chebotar, O.~Cortes, B.~David, C.~Finn, C.~Fu,
  K.~Gopalakrishnan, K.~Hausman, A.~Herzog, D.~Ho, J.~Hsu, J.~Ibarz, B.~Ichter,
  A.~Irpan, E.~Jang, R.~J. Ruano, K.~Jeffrey, S.~Jesmonth, N.~J. Joshi,
  R.~Julian, D.~Kalashnikov, Y.~Kuang, K.-H. Lee, S.~Levine, Y.~Lu, L.~Luu,
  C.~Parada, P.~Pastor, J.~Quiambao, K.~Rao, J.~Rettinghouse, D.~Reyes,
  P.~Sermanet, N.~Sievers, C.~Tan, A.~Toshev, V.~Vanhoucke, F.~Xia, T.~Xiao,
  P.~Xu, S.~Xu, M.~Yan, and A.~Zeng, ``Do {{As I Can}}, {{Not As I Say}}:
  {{Grounding Language}} in {{Robotic Affordances}},'' \emph{arXiv preprint
  arXiv:2204.01691}, 2022.

\bibitem{wang2025MOSAIC}
H.~Wang, K.~Kedia, J.~Ren, R.~Abdullah, A.~Bhardwaj, A.~Chao, K.~Y. Chen,
  N.~Chin, P.~Dan, X.~Fan, G.~{Gonzalez-Pumariega}, A.~Kompella, M.~A. Pace,
  Y.~Sharma, X.~Sun, N.~Sunkara, and S.~Choudhury, ``{{MOSAIC}}: {{Modular
  Foundation Models}} for {{Assistive}} and {{Interactive Cooking}},'' in
  \emph{CoRL}, 2024.

\bibitem{zuo2026Gaze2Act}
K.~Zuo, G.~Li, B.~Lyu, Y.~Lu, B.~Ma, S.~Han, X.~Zhou, X.~Yuan, C.~Zhou, J.~Bai,
  \emph{et~al.}, ``Gaze2act: Gaze-conditioned vision-language-action policies
  for interactive robot manipulation,'' \emph{arXiv preprint arXiv:2605.30282},
  2026.

\bibitem{solak2025Contextaware}
G.~Solak, G.~J.~G. Lahr, I.~Ozdamar, and A.~Ajoudani, ``Context-aware
  collaborative pushing of heavy objects using skeleton-based intention
  prediction,'' in \emph{ICRA}, 2025.

\bibitem{bhardwaj2021STORM}
M.~Bhardwaj, B.~Sundaralingam, A.~Mousavian, N.~Ratliff, D.~Fox, F.~Ramos, and
  B.~Boots, ``{{STORM}}: {{An Integrated Framework}} for {{Fast Joint-Space
  Model-Predictive Control}} for {{Reactive Manipulation}},'' in \emph{CoRL},
  2021.

\bibitem{ferraguti2020MulticameraSafety}
F.~Ferraguti, C.~{Talignani Landi}, S.~Costi, M.~Bonfè, S.~Farsoni, C.~Secchi,
  and C.~Fantuzzi, ``Safety barrier functions and multi-camera tracking for
  human–robot shared environment,'' \emph{Robotics and Autonomous Systems},
  vol. 124, p. 103388, 2020.

\bibitem{cirillo2012AmbientPlanning}
M.~Cirillo, L.~Karlsson, and A.~Saffiotti, ``Human-aware planning for robots
  embedded in ambient ecologies,'' \emph{Pervasive and Mobile Computing},
  vol.~8, no.~4, pp. 542--561, 2012, special Issue on Ambient Ecologies.

\bibitem{salzmann2020trajectron_simplified}
T.~Salzmann, B.~Ivanovic, P.~Chakravarty, and M.~Pavone, ``Trajectron++:
  Dynamically-feasible trajectory forecasting with heterogeneous data,'' in
  \emph{ECCV}, 2020.

\bibitem{bae2024singulartrajectory_simplifed}
I.~Bae, Y.-J. Park, and H.-G. Jeon, ``Singulartrajectory: Universal trajectory
  predictor using diffusion model,'' in \emph{CVPR}, 2024.

\bibitem{mao2020History}
W.~Mao, M.~Liu, and M.~Salzmann, ``History repeats itself: Human motion
  prediction via motion attention,'' in \emph{ECCV}, 2020.

\bibitem{manicast2023}
K.~Kedia, P.~Dan, A.~Bhardwaj, and S.~Choudhury, ``Manicast: Collaborative
  manipulation with cost-aware human forecasting,'' in \emph{Proceedings of The
  7th Conference on Robot Learning}, ser. Proceedings of Machine Learning
  Research, J.~Tan, M.~Toussaint, and K.~Darvish, Eds., vol. 229.\hskip 1em
  plus 0.5em minus 0.4em\relax PMLR, 06--09 Nov 2023, pp. 1050--1067.

\bibitem{kedia2024InteRACT}
K.~Kedia, A.~Bhardwaj, P.~Dan, and S.~Choudhury, ``{{InteRACT}}: {{Transformer
  Models}} for {{Human Intent Prediction Conditioned}} on {{Robot Actions}},''
  in \emph{ICRA}, 2024.

\bibitem{yolo26_ultralytics}
R.~Sapkota, R.~H. Cheppally, A.~Sharda, and M.~Karkee, ``Yolo26: key
  architectural enhancements and performance benchmarking for real-time object
  detection,'' \emph{arXiv preprint arXiv:2509.25164}, 2025.

\bibitem{jiang2023RTMPose}
T.~Jiang, P.~Lu, L.~Zhang, N.~Ma, R.~Han, C.~Lyu, Y.~Li, and K.~Chen,
  ``Rtmpose: Real-time multi-person pose estimation based on mmpose,''
  \emph{arXiv preprint arXiv:2303.07399}, 2023.

\bibitem{yoloe}
A.~Wang, L.~Liu, H.~Chen, Z.~Lin, J.~Han, and G.~Ding, ``Yoloe: Real-time
  seeing anything,'' in \emph{ICCV}, 2025.

\bibitem{li2026mv}
B.~Li, D.~Wu, J.~Li, S.~Zhou, Z.~Zeng, L.~Li, and H.~Zha, ``Mv-sam3d: Adaptive
  multi-view fusion for layout-aware 3d generation,'' \emph{arXiv preprint
  arXiv:2603.11633}, 2026.

\bibitem{wen2026fastfoundationstereo}
B.~Wen, S.~Dewan, and S.~Birchfield, ``{Fast-FoundationStereo}: Real-time
  zero-shot stereo matching,'' \emph{CVPR}, 2026.

\bibitem{liu2022rectifiedflow}
X.~Liu, C.~Gong, and Q.~Liu, ``Flow straight and fast: Learning to generate and
  transfer data with rectified flow,'' in \emph{ICLR}, 2023.

\bibitem{peebles2023scalable}
W.~Peebles and S.~Xie, ``Scalable diffusion models with transformers,'' in
  \emph{ICCV}, 2023.

\bibitem{rupprecht2017learning}
C.~Rupprecht, I.~Laina, R.~DiPietro, M.~Baust, F.~Tombari, N.~Navab, and G.~D.
  Hager, ``Learning in an uncertain world: Representing ambiguity through
  multiple hypotheses,'' in \emph{ICCV}, 2017.

\bibitem{chen2025dexonomy}
J.~Chen, Y.~Ke, L.~Peng, and H.~Wang, ``Dexonomy: Synthesizing all dexterous
  grasp types in a grasp taxonomy,'' \emph{Robotics: Science and Systems},
  2025.

\bibitem{sundaralingam2023curobo}
B.~Sundaralingam, S.~K.~S. Hari, A.~Fishman, C.~Garrett, K.~Van~Wyk, V.~Blukis,
  A.~Millane, H.~Oleynikova, A.~Handa, F.~Ramos, \emph{et~al.}, ``Curobo:
  Parallelized collision-free robot motion generation,'' in \emph{ICRA}, 2023.

\bibitem{wang2025moge2accuratemonoculargeometry}
\BIBentryALTinterwordspacing
R.~Wang, S.~Xu, Y.~Dong, Y.~Deng, J.~Xiang, Z.~Lv, G.~Sun, X.~Tong, and
  J.~Yang, ``Moge-2: Accurate monocular geometry with metric scale and sharp
  details,'' 2025. [Online]. Available: \url{https://arxiv.org/abs/2507.02546}
\BIBentrySTDinterwordspacing

\bibitem{nguyen2025gotrack}
V.~N. Nguyen, C.~Forster, S.~Shkodrani, V.~Lepetit, B.~Tekin, C.~Keskin, and
  T.~Hodan, ``Gotrack: Generic 6dof object pose refinement and tracking,''
  \emph{arXiv preprint arXiv:2506.07155}, 2025.

\bibitem{lin2025depthanything3}
H.~Lin, S.~Chen, J.~Liew, D.~Y. Chen, Z.~Li, G.~Shi, J.~Feng, and B.~Kang,
  ``Depth anything 3: Recovering the visual space from any views,'' \emph{arXiv
  preprint arXiv:2511.10647}, 2025.

\bibitem{labbe2022megapose}
Y.~Labb\'e, L.~Manuelli, A.~Mousavian, S.~Tyree, S.~Birchfield, J.~Tremblay,
  J.~Carpentier, M.~Aubry, D.~Fox, and J.~Sivic, ``Megapose: 6d pose estimation
  of novel objects via render \& compare,'' in \emph{Proceedings of the 6th
  Conference on Robot Learning (CoRL)}, 2022.

\bibitem{wen2025FoundationStereo}
B.~Wen, M.~Trepte, J.~Aribido, J.~Kautz, O.~Gallo, and S.~Birchfield,
  ``Foundationstereo: Zero-shot stereo matching,'' in \emph{Proceedings of the
  IEEE/CVF conference on computer vision and pattern recognition}, 2025, pp.
  5249--5260.

\bibitem{lin2024sam6d}
J.~Lin, L.~Liu, D.~Lu, and K.~Jia, ``Sam-6d: Segment anything model meets
  zero-shot 6d object pose estimation,'' in \emph{2024 IEEE/CVF Conference on
  Computer Vision and Pattern Recognition (CVPR)}.\hskip 1em plus 0.5em minus
  0.4em\relax IEEE, 2024, pp. 27\,906--27\,916.

\end{thebibliography}

\end{document}